%% file: sample-sigconf.tex
\documentclass[sigconf]{acmart}
\AtBeginDocument{%
  }

\AtBeginDocument{%
  }

\setcopyright{acmlicensed}
\copyrightyear{2018}
\acmYear{2018}
\acmDOI{XXXXXXX.XXXXXXX}
\acmISBN{978-1-4503-XXXX-X/2018/06}

\usepackage{multirow}

\begin{document}

\title{TabClaw: An Interactive and Self-Evolving Agent for Spreadsheet Manipulation and Table Reasoning}

\author{Mingyue Cheng}
\orcid{0000-0001-9873-7681}
\affiliation{%
  \institution{State Key Laboratory of Cognitive Intelligence, University of Science and Technology of China}
  \city{Hefei}
  \state{}
  \country{China}
}
\email{mycheng@ustc.edu.cn}

\author{Shuo Yu}
\orcid{0009-0006-1060-5451}
\affiliation{%
  \institution{State Key Laboratory of Cognitive Intelligence, University of Science and Technology of China}
  \city{Hefei}
  \state{}
  \country{China}
}
\email{yu12345@mail.ustc.edu.cn}

\author{Daoyu Wang}
\orcid{0009-0002-0452-0516}
\affiliation{%
  \institution{State Key Laboratory of Cognitive Intelligence, University of Science and Technology of China}
  \city{Hefei}
  \state{}
  \country{China}
}
\email{daoyu.wang@mail.ustc.edu.cn}

\author{Qingchuan Li}
\orcid{0009-0009-9747-0888}
\affiliation{%
  \institution{State Key Laboratory of Cognitive Intelligence, University of Science and Technology of China}
  \city{Hefei}
  \state{}
  \country{China}
}
\email{chouli@mail.ustc.edu.cn}

\author{Xiaoyu Tao}
\orcid{0009-0000-0634-6254}
\affiliation{%
  \institution{State Key Laboratory of Cognitive Intelligence, University of Science and Technology of China}
  \city{Hefei}
  \state{}
  \country{China}
}
\email{txytiny@mail.ustc.edu.cn}

\author{Qingyang Mao}
\orcid{0000-0002-6922-856X}
\affiliation{%
  \institution{State Key Laboratory of Cognitive Intelligence, University of Science and Technology of China}
  \city{Hefei}
  \state{}
  \country{China}
}
\email{maoqy0503@mail.ustc.edu.cn}

\author{Yitong Zhou}
\orcid{0009-0007-6579-1092}
\affiliation{%
  \institution{State Key Laboratory of Cognitive Intelligence, University of Science and Technology of China}
  \city{Hefei}
  \state{}
  \country{China}
}
\email{yitong.zhou@mail.ustc.edu.cn}

\author{Qi Liu}
\orcid{0000-0001-6956-5550}
\affiliation{%
  \institution{State Key Laboratory of Cognitive Intelligence, University of Science and Technology of China}
  \city{Hefei}
  \state{}
  \country{China}
}
\email{qiliuql@ustc.edu.cn}

\renewcommand{\shortauthors}{Mingyue Cheng et al.}
\begin{abstract}
\input{samples/sections/abstract}
\end{abstract}



\maketitle

\input{samples/sections/introduction}

\bibliographystyle{ACM-Reference-Format}
\bibliography{ref}

\end{document}

%% file: samples/sections/abstract.tex
Spreadsheets and tables are widely used representations for structured data analysis, but effective analysis still requires substantial manual effort and domain expertise. Recent large language model (LLM) agents can automate parts of this process, but they often provide limited transparency into intermediate decisions, rely on implicit assumptions, struggle with multi-table comparison, and repeat similar workflows without adapting to a user's preferences. This paper presents TabClaw, an open-source interactive AI agent for spreadsheet manipulation and table reasoning. Users upload CSV or Excel files and issue natural-language requests; TabClaw clarifies ambiguous intent, exposes an editable execution plan, streams a ReAct-style tool-using analysis loop, dispatches specialist agents for parallel multi-table reasoning, and synthesizes findings with explicit consensus and uncertainty markers. Beyond one-off analysis, TabClaw records completed workflows, extracts persistent user memory, distills reusable skills from repeated tool-use patterns, supports package-style skill import, and upgrades skills from negative feedback. Experiments on spreadsheet manipulation and table reasoning benchmarks show that TabClaw improves executable task completion and reasoning performance while preserving an inspectable user workflow. This paper shows how TabClaw turns spreadsheets and tables into inspectable analytical workflows while gradually personalizing itself to recurring data-analysis tasks. Our code is available.\footnote{ \url{https://github.com/ustc-table-mining/TabClaw}}

%% file: samples/sections/introduction.tex
\section{Introduction}

Spreadsheets and tables are widely used representations for structured data analysis. Analysts, researchers, and domain experts routinely ask questions over CSV files, spreadsheets, and exported database tables, but answering those questions still requires a chain of mechanical actions: inspect schemas, choose columns, clean missing values, filter rows, aggregate metrics, join related files, create pivot tables, and summarize the results. These operations are individually simple for experienced users, yet they become costly when users must repeatedly translate exploratory intent into spreadsheet formulas, SQL queries, or ad hoc scripts. The friction is especially visible in conversational settings, where a user may begin with an underspecified question such as ``which products are doing best?'', refine the target metric after seeing intermediate results, and then ask a follow-up comparison over another table.

Recent large language models (LLMs) have advanced table understanding and table reasoning~\cite{lu2025large,cheng2025survey}, making it possible to express analytical tasks in natural language. Prior work has explored table-specific pretraining~\cite{herzig2020tapas,wang2021tuta,liu2021tapex}, prompting and decomposition methods~\cite{jin2023tab,wang2024chain,ye2023large}, and LLM systems for spreadsheet or table manipulation~\cite{zhang2024tablellm,zha2023tablegpt,jiang2025tablemind}. These methods demonstrate strong progress, but many still treat table analysis as a one-shot reasoning problem: the model receives a table and a question, then returns an answer or a generated program. In practical data work, however, correctness depends not only on the final answer but also on whether the system supports user intervention and preserves useful experience across repeated workflows~\cite{cheng2026tablemind++,zhang2026star,wang2025tabletime}.

This gap motivates a more interactive view of table-analysis agents~\cite{schick2023toolformer,wang2024survey,shinn2023reflexion}. A useful agent should clarify ambiguous intent before acting, expose an editable plan, stream intermediate operations so that users can inspect and correct the process, and support multi-table analysis. It should also adapt to the user over time: recurring report patterns, domain-specific preferences, and negative feedback should become reusable system knowledge rather than disappearing after a single session. These requirements connect table reasoning with broader research on tool-augmented agents, planning, memory, and self-improvement~\cite{huang2024understanding,packer2023memgpt}.

In this paper, we present \textbf{TabClaw}, an open-source web system for interactive spreadsheet manipulation and table reasoning.\footnote{Demo video: \url{https://github.com/fishsure/TabClaw/blob/main/asset/TabClaw.mp4}.} TabClaw couples an LLM-driven ReAct loop~\cite{yao2023react,wang2026steppo} with a curated table-skill registry implemented on top of pandas. A user can upload one or more CSV/Excel files, ask a natural-language question, inspect an editable plan before execution, and watch the agent stream reasoning, tool calls, intermediate result tables, and final conclusions in real time. The system is designed for local deployment through a FastAPI backend and a browser-based interface, making it easy to demonstrate, extend, and audit.

TabClaw is designed to make table analysis both controllable during execution and reusable across sessions. When a request admits multiple plausible interpretations, the system asks clarification questions and allows the user to revise the plan before any data operation is executed. For comparative analysis over multiple uploaded tables, TabClaw assigns a scoped specialist agent to each table and then aggregates their findings with explicit indications of consensus and uncertainty. The system further maintains persistent memories about user preferences, domain facts, and recurring analysis patterns, and retrieves only relevant memories for future interactions. Completed workflows are also recorded so that repeated tool-use sequences can be distilled into reusable skills and improved when users provide negative feedback.

\begin{figure*}[t]
  \centering
  \includegraphics[width=\linewidth]{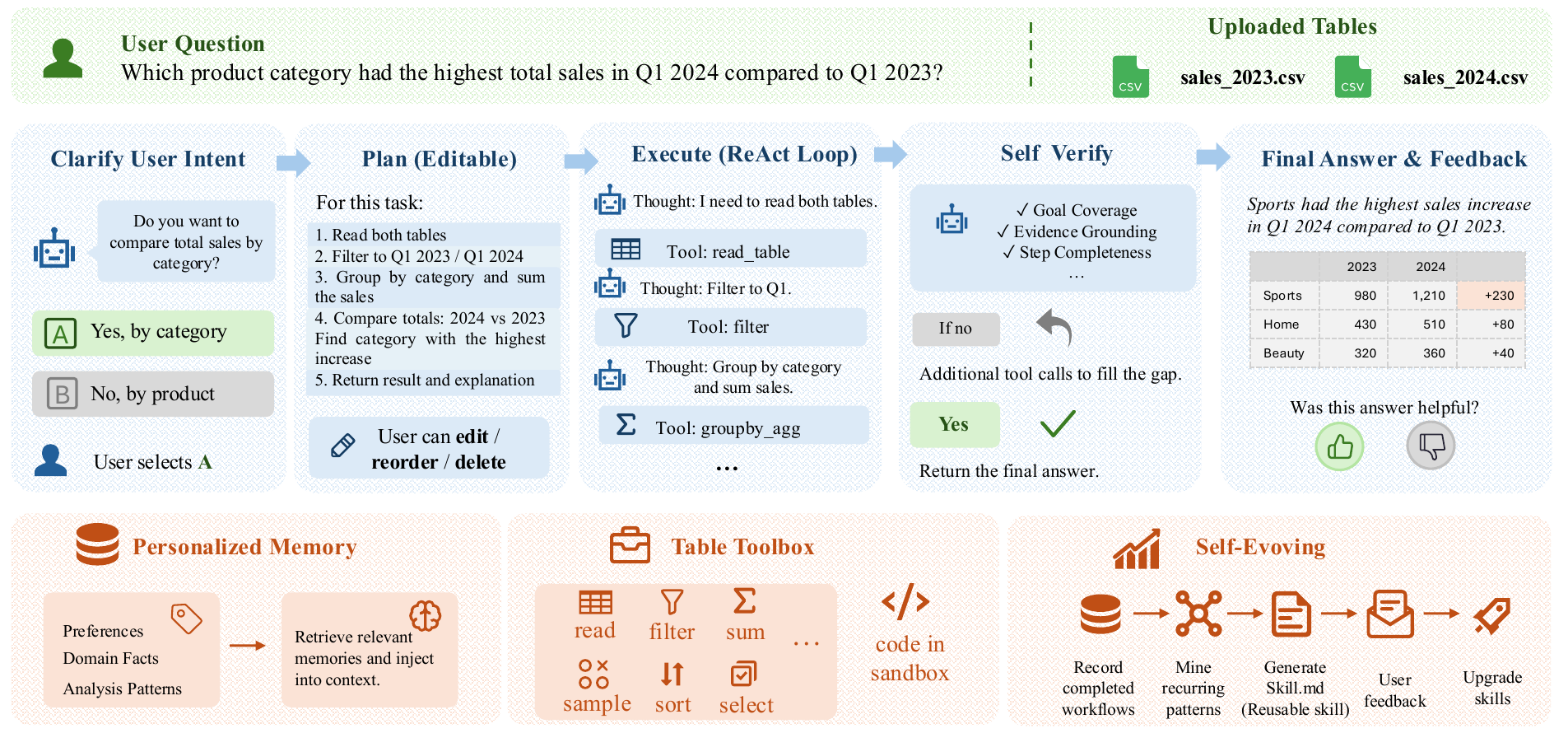}
  \caption{TabClaw workflow overview. The system turns a user question and uploaded tables into a clarified, planned, tool-grounded, and self-verified analysis, supported by personalized memory, a table toolbox, and workflow-driven skill evolution.}
  \label{fig:framework}
\end{figure*}

\section{System Overview}

Figure~\ref{fig:framework} summarizes the main workflow of TabClaw. The interaction begins with a user question and one or more uploaded tables. TabClaw first clarifies the user's analytical intent, then turns the request into an editable plan, executes the plan with tool-grounded reasoning, verifies whether the result is sufficiently supported, and finally returns an answer together with a feedback interface. This design makes the analysis process visible before, during, and after execution, so that users can intervene when the request is ambiguous, inspect intermediate evidence, and provide feedback that can improve future behavior.

The workflow is supported by three persistent components. Personalized memory stores user preferences, domain facts, and recurring analysis patterns, and retrieves relevant entries for future requests. The table toolbox provides structured operations such as reading, filtering, aggregation, sorting, sampling, selection, and sandboxed code execution. The self-evolving module records completed workflows, mines recurring patterns, distills reusable skills, and upgrades them based on user feedback. In implementation, the frontend is a single-page web interface for file upload, chat, plan editing, live event rendering, result-table inspection, memory management, and skill management, while the FastAPI backend exposes HTTP and Server-Sent Events (SSE) endpoints for the corresponding agent workflows.

\subsection{Intent Clarification and Planning}

Before any data operation is executed, TabClaw checks whether the user's request has multiple plausible interpretations. The clarification module considers the natural-language question together with the schemas of the uploaded tables, and asks a concise follow-up question only when different interpretations would lead to meaningfully different analyses. 
After the intent is resolved, TabClaw generates an editable plan containing concrete analysis steps, aligning with recent plan-then-execute and interactive planning paradigms~\cite{mao2024potable,erdogan2025plan,wang2023describe}. Users can inspect, reorder, delete, or rewrite steps before execution. This stage is important because spreadsheet tasks often require assumptions about columns, filters, grouping keys, and comparison baselines; exposing the plan gives users a chance to correct these assumptions before they affect the data.

\subsection{ReAct Execution and Self-Verification}

During execution, TabClaw follows a ReAct-style loop in which the model alternates between reasoning over the current task state and calling table tools. Each tool call produces an observation, such as a filtered table, an aggregate result, or a preview of relevant rows, and the next reasoning step is conditioned on these observations. The loop is streamed to the browser through SSE, allowing users to inspect the sequence of thoughts, tool calls, and intermediate result tables rather than receiving only an opaque final answer.
When a request involves multiple uploaded tables, TabClaw can assign scoped specialist agents to different tables and then aggregate their findings. Each specialist receives only the table and schema relevant to its subtask, which reduces cross-table contamination and makes intermediate conclusions easier to attribute. The final aggregation distinguishes high-confidence agreements from conflicts or caveats across tables.
After the planned steps finish, TabClaw runs a self-verification stage before returning the final answer. The verifier checks whether the original goal has been covered, whether the conclusion is grounded in observed evidence, and whether the planned steps have been completed. If the check identifies a gap, the system can issue additional tool calls and revise the answer; otherwise, it proceeds to the final response and collects user feedback.

\subsection{Persistent Support Modules}

The lower part of Figure~\ref{fig:framework} shows the modules that support the workflow across sessions. Personalized memory stores stable preferences, domain facts, and recurring analysis patterns, and retrieves relevant memories when a new request is processed. This allows TabClaw to reuse user-specific context without requiring the user to restate the same assumptions in every session.
The table toolbox provides the operations used by the ReAct executor. TabClaw currently includes 16 built-in pandas skills, including table inspection, filtering, column selection, aggregation, sorting, merging, pivoting, computed-column creation, descriptive statistics, value search, deduplication, renaming, sampling, value counts, correlation matrices, and head-row preview. For advanced cases, users can enable a code tool that runs sandboxed pandas and numpy code after AST inspection and timeout enforcement.
The self-evolving module connects completed workflows with future capability improvement. TabClaw records the sequence of tool calls, intermediate results, final conclusions, and user feedback for nontrivial analyses. Recurring successful patterns can be distilled into reusable skills, while negative feedback provides evidence for revising skills that repeatedly fail. The following section describes these personalization and evolution mechanisms in more detail.
\begin{figure}[t]
  \centering
  \includegraphics[width=\linewidth]{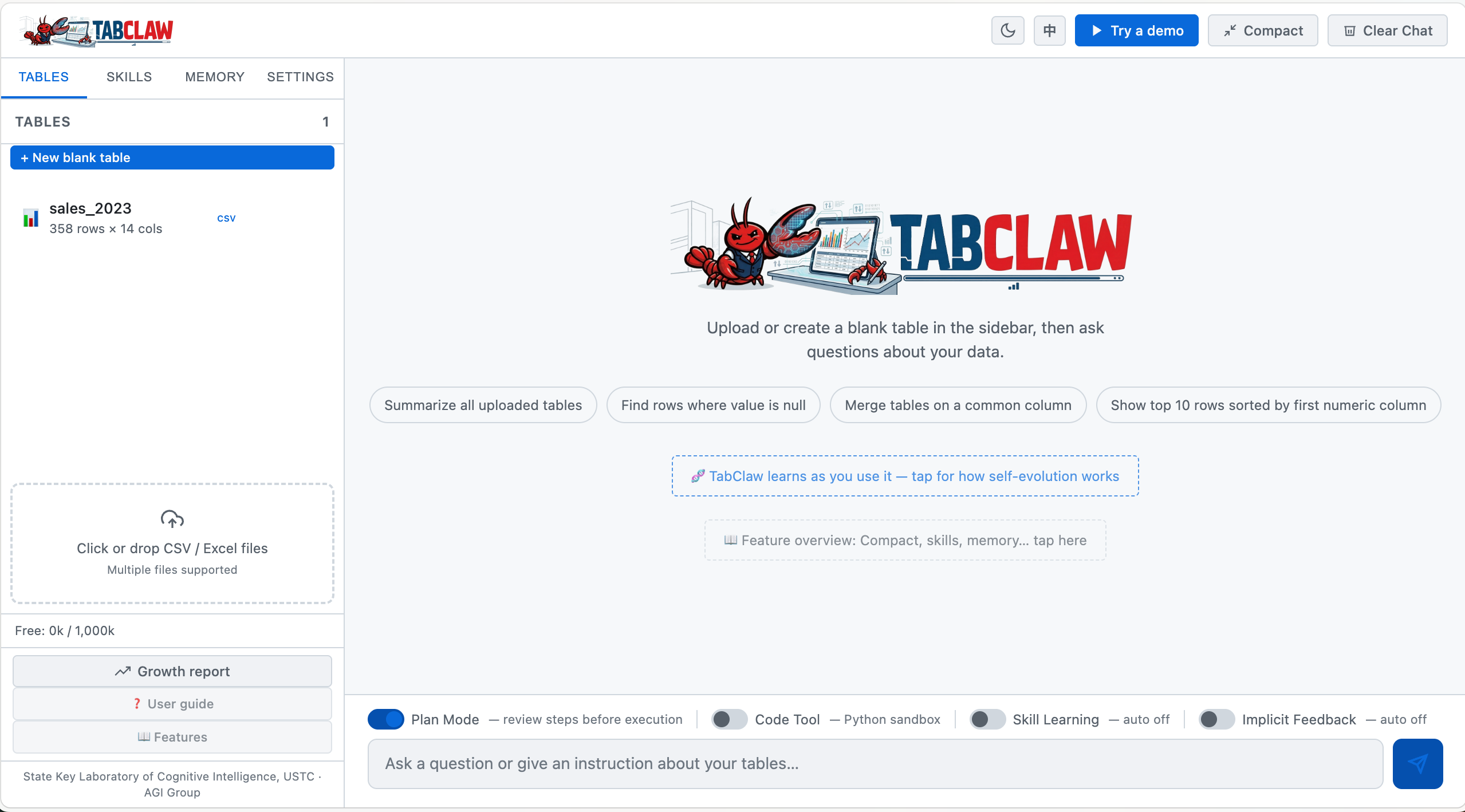}
  \caption{TabClaw browser interface with chat, streamed reasoning, table operations, and result inspection.}
  \label{fig:ui}
\end{figure}
\section{Self-Evolution and Personalization}
\input{samples/tables/performance}

TabClaw treats every completed nontrivial analysis as a workflow, storing the user request, involved tables, optional plan, ordered tool steps, intermediate summaries, final conclusion, feedback, and skills used. These records allow the system to reuse experience across sessions rather than treating each analysis as isolated. When a completed task contains a sufficiently rich tool sequence, TabClaw can assess whether the sequence captures a reusable pattern, such as a parameterized report, a top-$k$ analysis, a KPI summary, or a multi-step join, and distill it into a reusable skill for future conversations. Users can also trigger pattern discovery over historical workflows, where recurring tool sequences are grouped, compared with existing skills, and proposed as new skills before installation. Feedback further supports self-improvement: negative ratings are attached to both the workflow and the skills used in it, and repeated failures provide evidence for revising the corresponding skill, following recent work on self-correction and verbal feedback for language agents~\cite{gou2023critic,shinn2023reflexion}. Personalization is handled by persistent memory, a common design concern in long-running LLM agents~\cite{packer2023memgpt}. After each interaction, a lightweight extraction step identifies preferences, domain knowledge, user context, or historical insights worth remembering; at query time, stable preferences are included while other memories are retrieved only when relevant, keeping the prompt compact.

\section{Demonstration Scenario}

The demonstration is organized around three audience-operable scenarios in the browser interface shown in Figure~\ref{fig:ui}. First, in an ambiguity-aware analysis scenario, a user uploads a sales table and asks, ``Which products are doing best?'' TabClaw inspects the schema and asks whether ``best'' should mean revenue, profit, units sold, or growth. After the user selects a metric, the system generates an editable plan; the user can revise the grouping key, such as changing product-level analysis to category-level analysis, before execution. The audience then sees streamed tool calls, intermediate aggregates, and the final ranked answer.
Second, in the main multi-table comparison scenario aligned with Figure~\ref{fig:framework}, the user uploads 2023 and 2024 sales tables and asks which category has the highest Q1 growth. TabClaw assigns each table to a scoped specialist agent, computes table-specific Q1 summaries, and passes the results to an aggregator. The interface exposes specialist outputs side by side and then presents a synthesized answer with consensus, conflicts, and uncertainty markers, making clear how multi-table evidence is combined rather than hidden inside a single response.
Third, the demonstration shows self-evolution from feedback. After repeated tasks such as monthly KPI reports, top-$k$ growth reports, and anomaly summaries, TabClaw proposes a distilled \texttt{Skill.md} from historical workflows. The user reviews and approves installation. On a later similar request, the system retrieves the skill, executes a shorter and more stable tool sequence than the original ReAct exploration, and records feedback for future skill refinement.

\input{samples/sections/experiments}

\section{Conclusion}

This paper presented TabClaw, an interactive and self-evolving system for spreadsheet manipulation and table reasoning. TabClaw exposes table analysis as an inspectable workflow that combines clarification, editable planning, streamed tool execution, self-verification, memory, and skill evolution. The live demonstration shows how users can revise intermediate steps, compare multiple tables, and reuse recurring analysis patterns in a browser-based interface. Preliminary benchmark results further indicate performance gains over direct inference, context augmentation methods, and autonomous agent baselines. TabClaw provides a practical platform for studying transparent and adaptive human-agent collaboration on tabular data.

\section*{GenAI Usage Disclosure}

Generative AI tools were used to assist with language drafting and editing. The authors remain responsible for the technical content, claims, references, and final manuscript.

%% file: samples/tables/performance.tex
\begin{table*}[t]
\centering
\caption{Performance comparison on table and spreadsheet benchmarks.}
\label{tab:performance}
\begin{tabular}{llccccccc}
\toprule
\multirow{2}{*}{Category} & \multirow{2}{*}{Method} & \multicolumn{2}{c}{SpreadsheetBench} & \multicolumn{2}{c}{SheetCopilot} & \multirow{2}{*}{WikiTQ} & \multirow{2}{*}{TabFact} & \multirow{2}{*}{HiTab} \\
\cmidrule(lr){3-4} \cmidrule(lr){5-6}
 & & Soft Acc. & Hard Acc. & Exec@1 & Pass@1 & & & \\
\midrule
Direct Inference & Vanilla         & 0.188 & 0.155 & 0.955 & 0.694 & 0.614 & 0.734 & 0.636 \\
\midrule
\multirow{4}{*}{\shortstack[l]{Context \\ Augmentation}}
 & Binder         & \underline{0.241} & 0.189 & 0.965 & 0.693 & 0.725 & 0.854 & 0.658 \\
 & Chain-of-Table & 0.148 & 0.129 & 0.871 & 0.613 & 0.721 & 0.828 & 0.634 \\
 & NormTab        & 0.148 & 0.122 & 0.975 & \underline{0.728} & 0.659 & 0.757 & 0.603 \\
 & SpreadsheetLLM & 0.233 & \underline{0.198} & \underline{0.987} & 0.715 & 0.661 & 0.792 & 0.646 \\
\midrule
\multirow{2}{*}{\shortstack[l]{Autonomous \\ Agent}}
 & ReAct          & 0.188 & 0.161 & 0.968 & 0.726 & 0.698 & \underline{0.879} & 0.634 \\
 & SheetAgent     & 0.223 & 0.185 & 0.933 & 0.613 & \underline{0.732} & 0.868 & \underline{0.691} \\
\midrule
Ours & TabClaw        & \textbf{0.288} & \textbf{0.245} & \textbf{0.998} & \textbf{0.774} & \textbf{0.752} & \textbf{0.921} & \textbf{0.698} \\
\bottomrule
\end{tabular}
\end{table*}

%% file: samples/sections/experiments.tex
\section{Experiments}
\label{sec:experiments}

\subsection{Experimental Setup}

The evaluation covers SpreadsheetBench~\cite{ma2024spreadsheetbench} for spreadsheet manipulation and reasoning, SheetCopilot~\cite{li2023sheetcopilot} for executable spreadsheet operations, and WikiTQ~\cite{pasupat2015compositional}, TabFact~\cite{chen2019tabfact}, and HiTab~\cite{cheng2021hitab} for table question answering, fact verification, and hierarchical table reasoning. We report each benchmark's standard metric and compare against direct inference, context augmentation methods (Binder~\cite{cheng2022binding}, Chain-of-Table~\cite{wang2024chain}, NormTab~\cite{nahid2024normtab}, SpreadsheetLLM~\cite{dong2024spreadsheetllm}), and agent baselines (ReAct~\cite{yao2023react}, SheetAgent~\cite{chen2024sheetagent}). All methods use DeepSeek-v3.2~\cite{liu2025deepseek} as their backbone.

\subsection{Experimental Result}

Table~\ref{tab:performance} summarizes the results. TabClaw achieves the best performance across all reported benchmarks. On SpreadsheetBench, it improves both soft and hard accuracy, suggesting that explicit planning and tool-grounded execution are useful for operation-heavy spreadsheet tasks. On SheetCopilot, TabClaw also obtains the strongest execution success and pass-rate results, indicating that the system more reliably selects executable operations rather than only generating plausible textual answers.
The improvements also hold on WikiTQ, TabFact, and HiTab, which test compositional question answering, table-based fact verification, and hierarchical table reasoning, respectively. This consistency suggests that TabClaw's workflow benefits are not limited to a single table format or task family. We attribute the gains mainly to three factors: editable planning reduces premature commitment to an incorrect interpretation, the pandas-backed toolbox grounds reasoning in deterministic table operations, and streamed intermediate tables make the process inspectable and easier to correct. These results are still preliminary because the current evaluation focuses on task accuracy and execution success; future work should include human-centered measures such as time-to-answer, correction effort, user trust, and the long-term utility of memory and skill evolution.